\pdfoutput=1

\documentclass[11pt]{article}

\usepackage{EMNLP2023}

\usepackage{times}
\usepackage{latexsym}

\usepackage{amssymb}

\usepackage{graphicx}
\usepackage{subfigure}

\usepackage{color}
\usepackage{tabularray}

\usepackage[T1]{fontenc}

\usepackage[utf8]{inputenc}

\usepackage{microtype}

\usepackage{inconsolata}

\newcommand{\comment}[1]{}

\title{Disentangling the Linguistic Competence of Privacy-Preserving \texttt{BERT}}

\author{Stefan Arnold, Nils Kemmerzell \and Annika Schreiner \\ Friedrich-Alexander-Universität Erlangen-Nürnberg \\ Lange Gasse 20, 90403 Nürnberg, Germany \\ \texttt{(stefan.st.arnold, nils.kemmerzell, annika.schreiner)@fau.de}}

\begin{document}

\maketitle

\begin{abstract}

\textit{Differential Privacy} (DP) has been tailored to address the unique challenges of text-to-text privatization. However, text-to-text privatization is known for degrading the performance of language models when trained on perturbed text. Employing a series of interpretation techniques on the internal representations extracted from \texttt{BERT} trained on perturbed pre-text, we intend to disentangle at the linguistic level the distortion induced by differential privacy. Experimental results from a representational similarity analysis indicate that the overall similarity of internal representations is substantially reduced. Using probing tasks to unpack this dissimilarity, we find evidence that text-to-text privatization affects the linguistic competence across several formalisms, encoding localized properties of words while falling short at encoding the contextual relationships between spans of words.


\end{abstract}

\section{Introduction}

\textit{Language Models} (LM) \citep{devlin2018bert, radford2018improving} are among the most successful applications of machine learning and applied in a diverse range of tasks such as classification, translation, summarization, and question answering.

However, concerns were raised that LMs \citep{carlini2019secret, pan2020privacy} in general and their embedding layers \citep{song2020information, thomas2020investigating} in particular memorize and disclose personally identifiable information.

To mitigate the risk of information leakage due to unintended memorization, \textit{Differential Privacy} (DP) \citep{dwork2006calibrating} has been integrated into machine learning \citep{abadi2016deep} and LMs \citep{mccann2017learned,shi2022selective}. DP formalizes privacy through a notion of indistinguishability which is accomplished by injecting additive noise. 

While early adaptations of DP into LMs were applied to gradient updates \citep{mcmahan2017learning}, there is a shift towards applying DP on raw text \citep{fernandes2019generalised, feyisetan2020privacy, qu2021natural} in the form of text-to-text privatization. This technique aims to provide plausible deniability \citep{bindschaedler2017plausible} by perturbing words in a way that conceals authors and content.

\citet{qu2021natural} applied text-to-text privatization to \texttt{BERT} \citep{devlin2018bert} and explored techniques for privacy-adaptive pre-training (\textit{e.g.}, predicting a set of perturbed tokens for each masked position) and privacy-constrained fine-tuning. We complement this research direction by borrowing from range of techniques for model introspection to identify and localize the layer-wise alterations caused by perturbed text on internal representations and associate these with the retention and destruction of linguistic competence. 

Drawing on a representational similarity analysis \cite{kriegeskorte2008representational}, we measure a substantial dissimilarity between internal representations obtained from different privacy modalities. To connect this dissimilarity with linguistic formalisms, we conduct a series of probing tasks \citep{adi2016fine, tenney2019you, hewitt2019structural}. By contrasting the probing accuracies for recovering a range of twelve linguistic formalisms, we uncover that linguistic formalisms relying on localized properties endure the perturbations introduced by text-to-text privatization while properties that require context information are less resilient.

Since internal representations of LMs are formed by an attention mechanism \citep{vaswani2017attention}, we further investigate the distribution of attention patterns. By clustering the attention maps \citep{clark2019does}, we uncover that text-to-text privatization amplifies redundancy \citep{kovaleva2019revealing}.

\section{Preliminaries}
\label{sec:2}

\subsection{Language Models} 

Language Models (LMs) convert sentences composed of variable-length sequences of discrete tokens, such as \textit{characters}, \textit{subwords}, or \textit{words}, into fixed-length continuous embeddings. 

The introduction of the \textit{Transformer} architecture \cite{vaswani2017attention} and variants based solely on a encoder \cite{devlin2018bert} or decoder \cite{radford2019language} rapidly replaced recurrent architectures \cite{peters2018deep}. By relying entirely on a self-attention mechanism, transformers excel at modeling long-range interactions within text.

We focus on \texttt{BERT} \cite{devlin2018bert} with an uncased vocabulary, which exemplifies a family of transformers that produce bidirectional representations solely from the encoder block \cite{lan2019albert, sanh2019distilbert, liu2019roberta}.

The conventional workflow for \texttt{BERT} consists of two stages: \textit{pre-training} and \textit{fine-tuning}. During pre-training, \texttt{BERT} is trained on a pre-text corpus using masked language modeling (prediction of randomly masked words) and next sentence prediction (binarized prediction whether text pairs are adjacent). Fine-tuning involves adding a fully-connected layer trained end-to-end on labeled data, allowing \texttt{BERT} to adapt to various task related to language understanding \citep{wang2018glue}.

The internals of \texttt{BERT} comprise an embedding layer and multiple transformer layers. Once a text is tokenized into wordpieces \cite{wu2016google}, the embedding layer serves as a lookup table that contains a lexical representation for each token. Since \texttt{BERT} processes all token representations in parallel, the lexical representations need to be integrated with position and segment information. The transformer layers build on an attention mechanism that computes a scalar attention weight between each ordered pair of tokens and uses this weight to control the contextualization from every token regardless of its position or segment. Contextual representations together with attention maps provide the starting point for interpreting linguistic properties captured during pre-training \citep{tenney2019bert} and retained after fine-tuning \citep{merchant2020happens}.


\subsection{Differential Privacy}

Differential Privacy (DP) \citep{dwork2006calibrating} transitioned from the field of statistical databases into machine learning \citep{song2013stochastic, bassily2014private, abadi2016deep, shi2022selective}. DP operates on the principle of injecting additive noise so that model outputs are indistinguishable within the bounds of a privacy budget $\varepsilon > 0$, where $\varepsilon \rightarrow \infty$ represents no bound on the information leakage. 


Equipped with a discrete vocabulary set $\mathcal{W}$, an embedding function $\phi : \mathcal{W} \rightarrow \mathbb{R}$, and a distance metric $d: \mathbb{R} \times \mathbb{R} \rightarrow [0,\infty)$, \citet{feyisetan2020privacy} formulated a randomized mechanism for text-to-text privatization grounded in metric differential privacy \citep{chatzikokolakis2013broadening}. Specifically, the randomized mechanism perturbs each word in a text by adding noise to the representation of the word derived from an embedding space \citep{mikolov2013efficient} and projecting the noisy representation back to a discrete vocabulary using a nearest neighbor search. Since metric differential privacy scales the notion of indistinguishability by a distance $d(\cdot)$, this technique offers several benefits: (1) It ensures that the log-likelihood ratio of observing any substitution $\hat{w}$ given two words $w$ and $w’$ is bounded by $\varepsilon d\{\phi(w), \phi(w')\}$, providing plausible deniability \citep{bindschaedler2017plausible} with respect to all $w \in \mathcal{W}$. (2) It produces similar substitutions $\hat{w}$ for any words $w$ and $w'$ that are close in the embedding space, alleviating the curse of dimensionality associated with randomized response \citep{warner1965randomized}.

Table \ref{tab:example} illustrates an example output obtained by querying the randomized mechanism for text-to-text privatization. Notice that the fidelity to the original text is proportional to the privacy budget. However, the example also shows that text-to-text privatization suffers from many constraints such as grammatical errors \citep{mattern2022limits}, which spawned further developments aimed at improving both utility \citep{yue2021differential, arnold2023guiding, chen2023customized} and privacy \citep{xu2020differentially}.


\begin{table}
\centering
\caption{Example chunk (truncated) from \texttt{Wikipedia} privatized with different privacy budgets. Highlighted words represent a mismatch between the original word and the surrogate word after privatization.}
\label{tab:example}
\resizebox{\linewidth}{!}{%
\begin{tblr}{
  width = \linewidth,
  colspec = {Q[127]Q[815]},
  column{1} = {c},
  hline{1-2,4} = {-}{},
}
$\varepsilon$ & \textbf{Example}\\
$\infty$ & 'anarchism', 'is', 'a', 'political', 'philosophy', 'and', 'movement', 'that', 'is', 'skeptical', 'of', 'authority', 'and', 'rejects', 'all', 'involuntary', ',', 'coercive', 'forms', 'of', 'hierarchy', '.'\\

10 & '\textcolor{red}{syndicalism}', '\textcolor{red}{situated}', 'a', 'political', '\textcolor{red}{pedagogy}', '\textcolor{red}{but}', 'movement', 'that', '\textcolor{red}{help}', '\textcolor{red}{signalled}', '\textcolor{red}{the}', '\textcolor{red}{recommendation}', '\textcolor{red}{18}', 'rejects', '\textcolor{red}{four}', '\textcolor{red}{mobility}', ',', '\textcolor{red}{punitive}', 'forms', '\textcolor{red}{on}', '\textcolor{red}{associations}', '\textcolor{red}{outset}'
\end{tblr}
}
\end{table}


\comment{
\begin{equation}
\label{equation:dp2}
 \frac{\mathbb{P} [\mathcal{M}(w) = \hat{w}]}{\mathbb{P} [\mathcal{M}(w^{'}) = \hat{w}]} \leq e^{\varepsilon d\{\phi(w),\phi(w^{'})\}}.
\end{equation}
}

\subsection{Model Introspection}

Aimed at understanding the internals of language models, numerous interpretation techniques were developed to uncover which properties of a text are embedded in contextual representations. Prominent techniques include stimuli and diagnostic models.

\paragraph{Stimuli-based Probes.} \citet{linzen2016assessing} assembled texts containing curated stimuli and evaluated the perplexity scores on masked stimuli as evidence for the presence or absence of linguistic knowledge. Using a fill-mask objective on stimuli was adopted to examine a range of linguistic properties, in particular \textit{subject-verb agreement} \cite{gulordava2018colorless, marvin2018targeted, lakretz2019emergence, goldberg2019assessing, ettinger2020bert}. 


\paragraph{Classifier-based Probes.} \citet{adi2016fine} eliminated the need for curating stimuli by setting up probing models. A probing model inputs internal representations as features annotated by linguistic properties of interest as labels and its accuracy score is directly interpreted as the extent to which linguistic properties are contained in the internal representation. Since probing models require few assumptions beyond the existence of model activations, they are widely used to assess the linguistic competence of language models \cite{belinkov2017neural, conneau2018you, hupkes2018visualisation}.


Considerable research is centered on the inspection of fixed-length sentence representations. \citet{adi2016fine} introduced a probing suite to extract surface properties of sentences such as \textit{length}, \textit{content}, and \textit{order}. \citet{conneau2018you} later recasted and extend these probing tasks by a broader set of linguistic properties, such as \textit{tense} and \textit{depth}.

Contrary to probing fixed-length sentence representations, probing suits exist that are tailored towards linguistic properties in word-level representations \cite{blevins2018deep, peters2018dissecting, tenney2019you, liu2019linguistic}. \citet{tenney2019you} present \textit{edge probing} in which a diagnostic model is given access only to span representations. From these span representations, the probing model aims to extract high-level linguistic properties which are expected to require complete sentence context. The analysis of intermediate layers of language models indicates that linguistic properties are captured in a hierarchical order \cite{peters2018dissecting, tenney2019bert, jawahar2019does}. This hierarchy is composed of signals ranging from surface abstractions in the lower layers, syntactic abstractions in the middle layers and semantic abstractions in the higher layers.



While prior probes on detecting syntactic structure lacked an explanation of whether structure is embedded as an entire parse tree \cite{conneau2018you} or how such parse trees are embedded \cite{peters2018dissecting}, \citet{hewitt2019structural} proposed a \textit{structural probe} to recover the topology of an entire parse tree and derive its parse depth. Using a linear transformation of the representation space, the structural probe shows evidence of a geometric representation that implicitly embeds sentence structure. The structural hypothesis formed by the linear transformation has recently been refined by a scaled isomorphic rotation \cite{limisiewicz2020introducing}, kernelization using a radial-basis function \cite{white2021non}, and projection onto hyperbolic space \cite{chen2021probing}.


To examine how contextual representations are formed through the attention mechanism \citep{vaswani2017attention}, recent research extended their analysis to role of attention in handling properties of text \cite{lin2019open, jo2020roles}. The visualization of attention heatmaps and the calculation of the distribution of attention revealed interpretable positional patterns \cite{vig2019analyzing, clark2019does, kovaleva2019revealing} and strong correlations to linguistic properties \cite{clark2019does, htut2019attention, ravishankar2021attention}. 


\paragraph{Limitations.} Despite its popularity for model introspection, recent studies observed that linguistic properties are incidentally captured even without task relevance \citep{ravichander2020probing}, casting doubt on the interpretations derived from attention maps \citep{jain2019attention, serrano2019attention, brunner2019identifiability} and probing models \citep{tamkin2020investigating}. This prompted the design of control tasks \cite{hewitt2019designing, ravichander2020probing}, amnesic probing \cite{elazar2021amnesic, jacovi2021contrastive}, conditional probing \cite{hewitt2021conditional}, and orthogonal techniques for correlating contextual representations \cite{saphra2018understanding, voita2019bottom, abdou2019higher}.

\section{Methodology}
\label{sec:3}

We follow the convention of denoting words and sentences using italic $(\mathit{w_i},\mathit{s})$, and refer to their representations using bold $(\mathbf{w_i},\mathbf{s})$, where the index $i$ distinguishes words in a sentence. Let $d$ be the dimension of a $l$-layer LM. Given a sentence $\mathit{s}$ as a tokenized list of words $\mathit{w} \in \mathcal{W}$, the LM inputs a lexical vector representation for each word and computes a contextual vector representation $\mathbf{w}_i^l \in \mathbb{R}^d$ for the $i$-th word at the $l$-th layer. 

We pre-train \texttt{BERT} models from-scratch following \citet{devlin2018bert} on a dump of \texttt{Wikipedia} preprocessed with a privacy budget of $\epsilon \in \{10,\infty\}$, where $10$ yields a privacy-preserving \texttt{BERT} and $\infty$ serves as our baseline for comparison. Apart from the difference in the privacy modality, training is identical to erase any confounding factors. 

Equipped with \texttt{BERT} pre-trained on a corpus of \texttt{Wikipedia} with different privacy modalities, we intend to uncover how and where contextual representations produced by the model trained with differential privacy depart from those produced by the model trained without differential privacy. Following the experimental setup of \citet{merchant2020happens}, we address this question mainly through the lens of (unsupervised) representational similarity analysis and (supervised) probing models.



\subsection{Similarity Analysis} 

We aim to compare the internals of language models that originate from pre-training under public and private training environments. Due to the lack of correspondence between activation patterns of models trained with different modalities, we need to abstract away from direct comparison of model activations. We instead leverage \textit{Representational Similarity Analysis} (RSA) \cite{kriegeskorte2008representational} to correlate the dissimilarity structure between contextual representations. Building on dissimilarity structures rather than activation patterns, RSA is indifferent to the representation space.

We base our similarity analysis on higher-order comparisons introduced by \citet{abdou2019higher}. Given a set of language models trained under different (privacy) modalities $M$ and a common set of sentences $N$, we extract representations as layer-wise activations from each $M$. Using any kernel that satisfies the axioms of a (dis)similarity metric, we can convert the extracted representations into pairwise dissimilarity matrices $\mathbb{R}^{n \times n}$. Each $N \times N$ dissimilarity matrix corresponds to the dissimilarity between the activation patterns associated with sentences pairs $n_i, n_j \in N$. Since the dissimilarity is intuitively zero when $n_i = n_j$, the dissimilarity matrix is symmetric along a diagonal. Using another kernel, we can now correlate the similarity between the flattened upper triangulars of the constructed dissimilarity matrices. 

We adopt the \textit{Cosine distance} as metric for the \textit{intra}-space dissimilarity and \textit{Spearman correlation} as metric for the \textit{cross}-space similarity. The RSA is performed on a random subset of $5,000$ sentences drawn from \texttt{WikiText} \citep{merity2016pointer}.

\subsection{Linguistic Probing} 

We aim to connect the dissimilarity between contextual representations with linguistic properties. To discern and locate the extent to which linguistic properties of texts are captured, we employ probing tasks at word-level and sentence-level representations for a range of surface, syntactic, and semantic formalisms. Note that \texttt{BERT} uses tokenization into subwords. Since word-level probes require access to word representations, we map subword representations to word representations by element-wise mean pooling over all subword components.

\paragraph{Surface Probe.} We evaluate surface properties using the setup for sentence-level probing assembled by \citet{adi2016fine}. To form sentence representations $\mathbf{s} \in \mathbb{R}^{d}$, we use element-wise mean pooling. Without access to a sentence $\mathit{s}$ and any of its words $\mathit{w}$, the surface proprieties to extract are \textit{length}, \textit{content}, and \textit{order}. The length task measures to what extent a sentence representation $\mathbf{s}$ encodes the length $|\mathit{s}|$ of a sentence $\mathit{s}$. The length task is formulated as a multi-class classification for a balanced set of binned lengths in intervals $[0,35)$, $[35,41)$, $[41,46)$, $[46, 52)$, $[52, \infty)$. The content task measures the extent to which a sentence representation $\mathbf{s}$ encodes the identities of words $\mathit{w}$ in a sentence. The content task is formulated as a binary classification in the form $(\mathbf{s},\mathbf{w}) \in \{0,1\}$, where $0$ denotes $\mathit{w} \not\in \mathit{s}$ and $1$ denotes $\mathit{w} \in \mathit{s}$, respectively. The order task measures the extent to which a sentence representation $\mathbf{s}$ encodes the order of words $\mathit{w_i}$, $\mathit{w_j}$. Given a sentence representation $\mathbf{s}$ and two word representation $\mathbf{w_i}$, $\mathbf{w_j}$ of words appearing in a sentence, the content task is formulated as a binary classification in the form $(\mathbf{s},\mathbf{w_i},\mathbf{w_j}) \in \{0,1\}$, where $0$ denotes $\mathbf{w_i} \prec \mathbf{w_j}$ and $1$ denotes $\mathbf{w_i} \succ \mathbf{w_j}$, respectively. All surface probes are performed on sentences from the training set reflecting their presumably most accurate representations. 

\paragraph{Linguistic Probe.} To evaluate linguistic properties , we employ \textit{edge probes} \cite{tenney2019you} and \textit{structural probes} \cite{hewitt2019structural} as two complementary probes at word-level.

The purpose of edge probing is to measure the extent to which contextual representations capture syntactic dependencies and semantic abstractions. Instead of supplying a probing model with a pooled sentence representation $\mathbf{s}$, edge probing decomposes the probing task into a common format so that the probing model only receives labeled spans $[\mathbf{w}_{i}^{l},\mathbf{w}_{j}^{l})$ and (optionally) $[\mathbf{w}_{u}^{l},\mathbf{w}_{v}^{l})$. With access only to contextual representations within the end-exclusive spans, the probing model must label the relation between these spans and their role in the sentence. Derived from evaluation on tagged benchmark datasets, we report the micro-averaged harmonic mean of the precision and recall for labeling \textit{part-of-speech tags}, \textit{constituency phrases}, \textit{dependency relations} as syntactic tasks, and \textit{entity types}, \textit{entity relations}, \textit{semantic roles}, and \textit{coreference mentions} as semantic tasks.

The structural probe is designed to measure the representation of syntactic structure. The probe identifies whether the geometric space under linear transformation $B \in \mathbb{R}^{k \times d}$, where $k$ is the rank of the transformation and $d$ is the dimensionality of the representation, captures the depth of words or distances between words in a parse tree. We adjust the rank to the dimensionality $k=d$. The \textit{depth} probe measures the distance from root $\forall i$ in a parse tree. It is defined by $\|\textbf{w}_{i}^{l}\|_{B}=(B\textbf{w}_{i}^{l})^{T}(B\textbf{w}_{i}^{l})$. The depth probe is evaluated based on the accuracy of the root word and the correlation between the predicted order of words and ordering specified by the depth in the parse tree. The \textit{distance} probe measures the pairwise distances $\forall i,j$ within a parse tree. It is defined by $\|\textbf{w}_{i}^{l}-\textbf{w}_{j}^{l}\|_{B}=(B(\textbf{w}_{i}^{l}-\textbf{w}_{j}^{l})^{T}(B(\textbf{w}_{i}^{l}-\textbf{w}_{j}^{l}))$. The distance probe is evaluated by correlating the predicted distances between pairs of words with distances metrics specified by the parse tree and by converting the predicted distances between pairs of words into a minimum spanning tree and scoring it against the parse tree using the Undirected Unlabeled Attachment Score (UUAS).  

\section{Experiments}
\label{sec:4}

\begin{figure}[t]
    \includegraphics[width=0.45\textwidth]{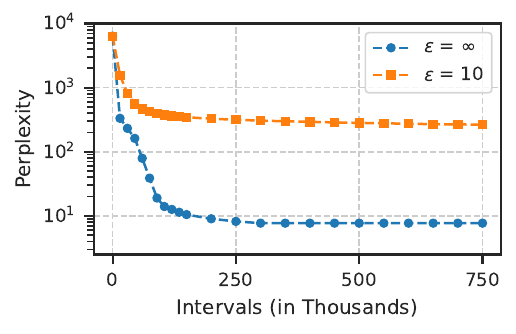}
    \caption{
      Interval-wise learning progress of \texttt{BERT} from $26,903,298$ chunks generated from \texttt{Wikipedia}. 
    }
    \label{fig:ppl}
\end{figure}

We initiate our model introspection by examining the performance in terms of perplexity scores. Figure \ref{fig:ppl} reveals that \texttt{BERT} trained on a corpus of text subjected to text-to-text privatization converges to a notably (but reasonably) worse perplexity score at $61.45$ (compared to $6.82$). Since perplexity is a measure for assessing the proficiency of language models in predicting the next word in a sentence, the elevated value in this context connotes a diminished ability for language modeling. To elucidate the linguistic alterations that lead to the degradation of the perplexity score, we pursue a layer-wise ablation of linguistic properties captured in the internal representations of privacy-preserving \texttt{BERT}.


\subsection{Similarity Results}

In line with correlation coefficients, RSA scores have value range of $[-1,+1]$, where $+1$ indicates that the models produce a similar internal representation and $-1$ indicates that the models diametrically opposed in latent space. Since these theoretical bounds are unlikely in practice, we establish an empirical bound on RSA by correlating the dissimilarity structures of \texttt{BERT} models with identical architecture but different initialization. We observe that the average similarity bounds at $0.9051$. By correlating the dissimilarity structures between \texttt{BERT} and \texttt{BERT} trained on perturbed text, we find a remarkable drop to $0.7601$, signifying a substantial departure between their internal representations.

\begin{figure}[t]
    \includegraphics[width=0.45\textwidth]{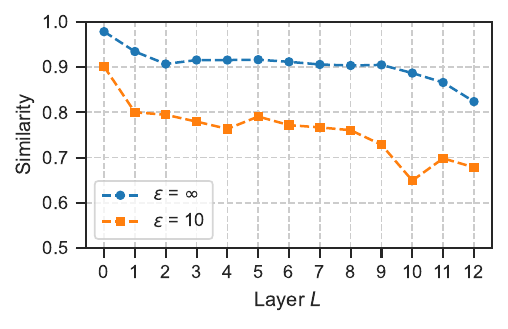}
    \caption{
     Layer-wise representational similarity of \texttt{BERT} for $5,000$ samples randomly drawn from \texttt{WikiText}.
    }
    \label{fig:rsa}
\end{figure}

\begin{figure*}[t]
    \subfigure[Text Length]
    {
        \centering
        \includegraphics[width=0.23\textwidth]{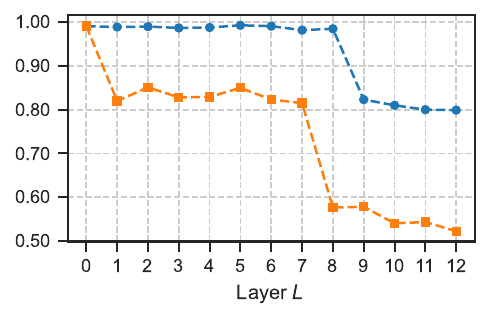}
        \label{fig:surface_length}
    }
    \subfigure[Word Content]
    {
        \centering
        \includegraphics[width=0.23\textwidth]{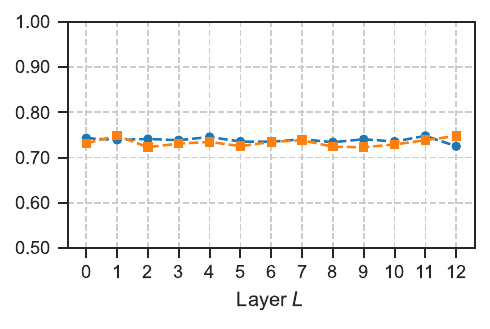}
        \label{fig:surface_content}
    }
    \subfigure[Word Order]
    {
        \centering
        \includegraphics[width=0.23\textwidth]{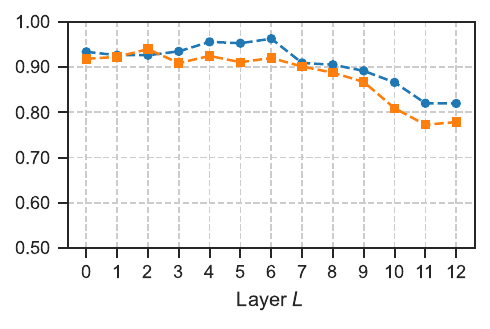}
        \label{fig:surface_order}
    }
    \newline
    \subfigure[Grammatical Tags]
    {
        \centering
        \includegraphics[width=0.23\textwidth]{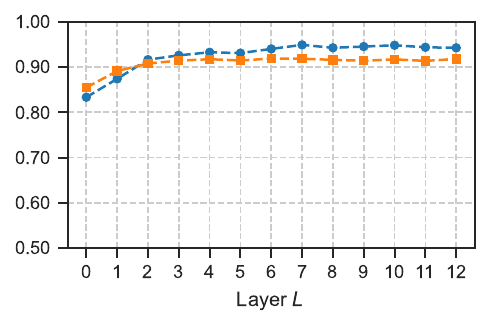}
        \label{fig:syntax_pos}
    }
    \subfigure[Constituency Chunks]
    {
        \centering
        \includegraphics[width=0.23\textwidth]{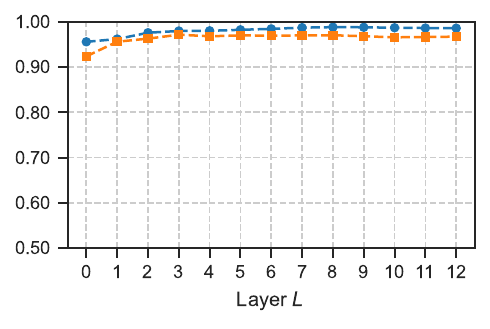}
        \label{fig:syntax_con}
    }
    \subfigure[Dependency Relations]
    {
        \centering
        \includegraphics[width=0.23\textwidth]{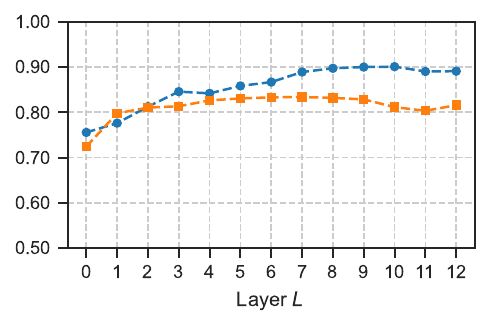}
        \label{fig:syntax_dep}
    }
    \newline
    \subfigure[Entity Types]
    {
        \centering
        \includegraphics[width=0.23\textwidth]{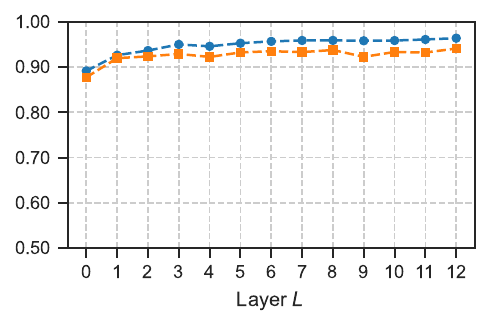}
        \label{fig:semantic_ent}
    }
    \subfigure[Entity Relations]
    {
        \centering
        \includegraphics[width=0.23\textwidth]{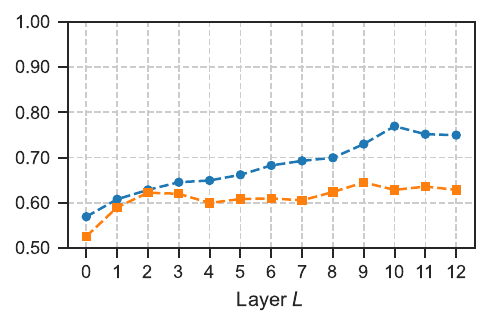}
        \label{fig:semantic_rel}
    }
    \subfigure[Semantic Roles]
    {
        \centering
        \includegraphics[width=0.23\textwidth]{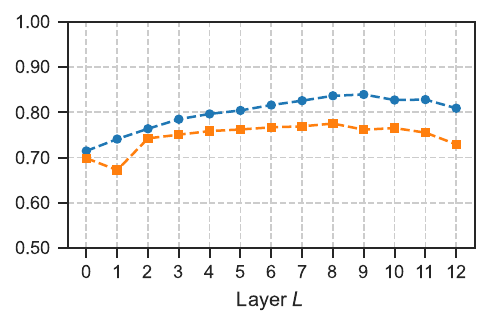}
        \label{fig:semantic_rol}
    }
    \subfigure[Coreference Mentions]
    {
        \centering
        \includegraphics[width=0.23\textwidth]{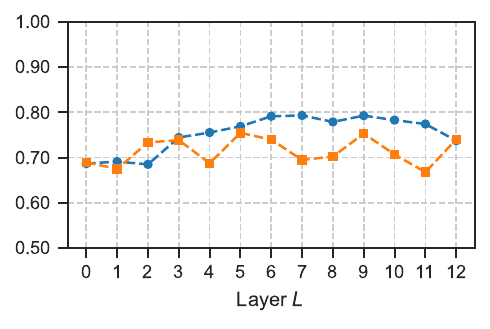}
        \label{fig:semantic_ref}
    }
    \newline
    \subfigure[Parse Depth]
    {
        \centering
        \includegraphics[width=0.23\textwidth]{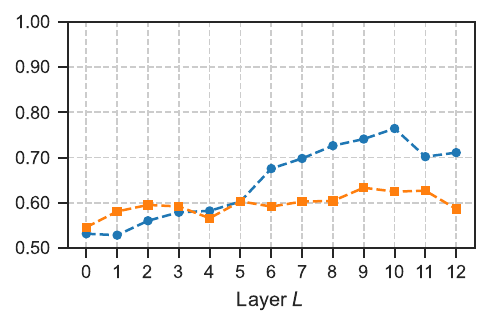}
        \label{fig:structure_depth}
    }
    \subfigure[Parse Distance]
    {
        \centering
        \includegraphics[width=0.23\textwidth]{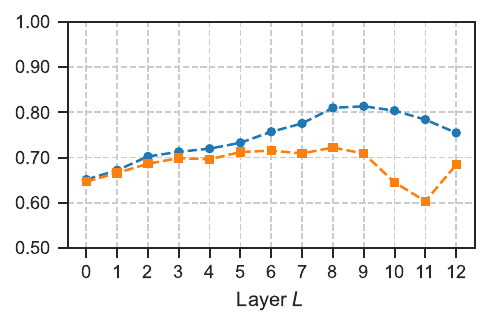}
        \label{fig:structure_distance}
    }
    \caption{
    Layer-wise probing results for \texttt{BERT} under public (blue circles) and private (orange squares) training modalities. Surface properties according to \citet{adi2016fine} are depicted in Figures \ref{fig:surface_length}, \ref{fig:surface_content}, and \ref{fig:surface_order}. Syntactic properties according to \citet{tenney2019you} are depicted in Figures \ref{fig:syntax_pos}, \ref{fig:syntax_con}, and \ref{fig:syntax_dep}. Semantic properties according to \citet{tenney2019you} are depicted in Figures \ref{fig:semantic_ent}, \ref{fig:semantic_rel}, \ref{fig:semantic_rol}, and \ref{fig:semantic_ref}. Structural properties according to \citet{hewitt2019structural} are depicted in Figures \ref{fig:structure_depth} and \ref{fig:structure_distance}.
    }
    \label{fig:probing}
\end{figure*}

To locate the variations in the internal representations on different layers of the \texttt{BERT} architecture, we present the layer-wise RSA results in Figure \ref{fig:rsa}. Note that \texttt{BERT} models typically maintain consistently high RSA values across all layers, whereas our \texttt{BERT} model trained on perturbed text starts with relatively high RSA values at the lexical representation layer at $0.9007$ and declines with contextual representations layers to $0.6784$, indicating a sharper deviation in the representation space. This pattern carries significant implications for our understanding of the impact of text-to-text privatization. Since the lexical representation corresponds to occurrence characteristics, this indicates that private \texttt{BERT} fails to capture context information.



\subsection{Probing Results}

Assuming that the substantial divergence arises from the fact that privacy-preserving \texttt{BERT} forms its contextual representation based on different linguistic properties than \texttt{BERT}, we are interested in discovering which linguistic properties are captured despite being trained on perturbed text. 

Figure \ref{fig:probing} depicts the probing results. The layer-wise probing results are shaped similarly but the consistently lower scores across all properties indicate that the linguistic competence is compromised when text-to-text privatization is are applied.

\paragraph{Surface.} Starting from the sentence-level probes, we notice distinct patterns in the details captured about surface properties. With a deficit of $-0.2770$, there is a marked difference related to the encoded text length. Contrasting this deficiency, details concerning content and order show a higher degree of consistency, reflecting deviations of $+0.0230$ and $-0.0410$, respectively. To grasp the implications of surface properties, we recall the argumentation of \citet{adi2016fine} that representations containing information about length and order are more suited for syntactic tasks while representations that excel at content are more suited for semantic tasks. 


\paragraph{Linguistic.} We continue with linguistic properties at word-level. From syntactic probes, we observe that a significant portion of information about grammatical tags and constituency chunks are retained at $-0.0246$ and $-0.0187$, while less emphasis is placed on capturing dependency relations, resulting in a reduction of $-0.0751$. From semantic probes, we notice that information about entity types is missing by only $-0.0229$, while entity relations and semantic roles experience a more substantial drop of $-0.1209$ and $-0.0798$. From structural probes, which test whether a representation encodes topology, we consolidate the findings from the syntactic probe on dependency relations. Scored against a discrete solution in the form of the root word or minimum spanning tree, the representations contain information about the root word with a score of $0.5866$ and the parse tree with a score of $0.6843$, representing decrements of $-0.1244$ and $-0.0703$, respectively.


Considering the nature of the linguistic properties and the degree to which they decline under privacy constraints, it is noticeable that formalisms closely related to basic characteristics of words display a considerable degree of preservation, whereas formalisms tied to complex relationships within spans of words undergo a substantial degree of deterioration. This intriguing pattern suggests that while localized properties endure the perturbations of text-to-text privatization, the ability of language models to maintain contextual constructs can be severely hindered by text-to-text privatization.

Since text-to-text privatization builds on word-level differential privacy \citep{mattern2022limits}, a plausible explanation for this phenomenon could be rooted in the nature of its randomized mechanism, which has been observed to disproportionately affect linguistic properties \citep{arnold2023guiding}. This insight underscores the interplay between perturbation strategies and the necessity of accurately conveying different types of linguistic formalisms.

\paragraph{Attention.} Since contextual representations are mainly formed by the mechanism of self-attention \citep{vaswani2017attention}, we could attribute the alterations in the representations to the fact that the attention mechanism (somehow) fails to discriminate certain linguistic properties. We attempt to answer this hypothesis by analyzing the distributional patterns of attention maps.

\begin{figure}[t]
    \centering
    \subfigure[\texttt{BERT} with DP at $\epsilon=\infty$]
    {
        \centering
        \includegraphics[width=.45\textwidth]{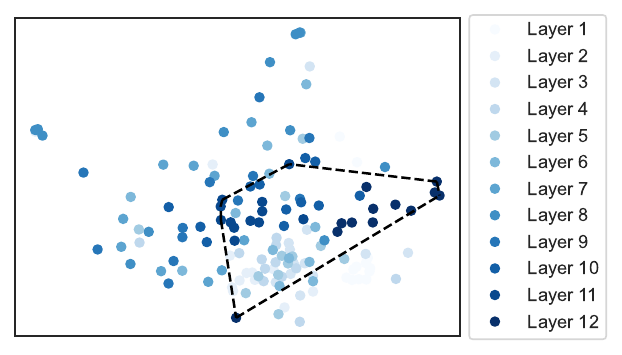}
        \label{fig:attention_cluster_base}
    }
    \hfill
    \subfigure[\texttt{BERT} with DP at $\epsilon=10$]
    {
        \centering
        \includegraphics[width=.45\textwidth]{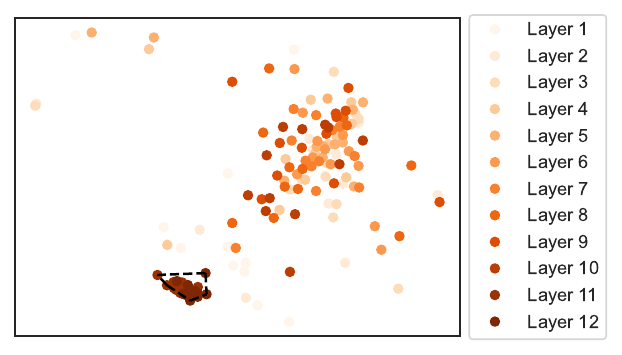}
        \label{fig:attention_cluster_priv}
    }
    \caption{Divergence-based clustering of attention maps extracted from $1,000$ random samples of \texttt{WikiText}.}
    \label{fig:attention_cluster}
\end{figure}

Once for each training modality, we obtain attention maps for $1,000$ randomly selected sentences and rearrange the attention maps from their subwords in line with \citet{vig2019analyzing}. For attentions drawn to a split-up word, we sum up the attention weights over its subwords. For attentions stemming from a split-up word, we average all weights from its subwords. Following \citet{clark2019does}, we calculate the distance between all pairs of attention maps using the Janson-Shannon divergence and visualize the distances grouped by layer using multidimensional scaling in Figure \ref{fig:attention_cluster}.

Assuming that attention heads that are clustered closely together perform similar linguistic roles in forming the internal representation, we conclude from the distributional patterns that text-to-text privatization amplifies the redundancy that is already present in attention heads as revealed by \citet{kovaleva2019revealing}. This is most evident by comparing the overlap of the attention maps in rear layers.

Considering that \citet{li2018multi} showed that encouraging the attention mechanism to have diverse behaviors can improve performance, we find another possible explanation for the lack of linguistic competence in privacy-preserving language models and their deteriorated level of perplexity. 


\section{Conclusion}
\label{sec:5}

Assuming that the performance loss of language models caused by text-to-text privatization can be attributed to the destruction of linguistic competence \citep{merendi2022nature}, we set to disentangle the layer-wise alterations of perturbations to the internal representations of a language model. 

By employing a series of techniques for model introspection \citep{adi2016fine, hewitt2019structural, tenney2019you}, we tested the internal representations formed by language models for linguistics properties across several formalisms.

From the perspective of linguistic competence, experimental results from our layer-wise model introspection indicate that privacy preservation can considered conservative as language models subjected to text-to-text privatization retain a hierarchical order of linguistic formalisms \citep{peters2018dissecting, tenney2019bert, jawahar2019does}. However, text-to-text privatization shows to have a cumulative impact on the linguistic competence of language models, affecting aspects ranging from surface-level properties to linguistic constructs across syntactic, semantic, and structural formalisms. We further notice that basic properties of words are less disrupted than complex relations between words that require context information.

\paragraph{Limitations.} Most assumptions and findings of this study are grounded in probing. Although probing enjoys much support as a technique for interpreting the internals of language models \citep{abadi2016deep,conneau2018you,tenney2019you,hewitt2019structural}, recent studies dispute with conclusion derived from probing due to the fact that probing may not entail task relevance \citep{ravichander2020probing}. We side with those viewing probing as a tool for model introspection, but nonetheless caution that our probing results may not be the appropriate technique for discerning the differences of private training modalities. Given the wide range of probing tasks and the fact that our probing results show a consistent pattern of competencies, we are convinced that this study contributes novel privacy implications.

\bibliography{anthology,custom}
\bibliographystyle{acl_natbib}

\appendix


\end{document}